\documentclass[conference]{IEEEtran} 
\usepackage{graphicx}
\usepackage{rotating}
\usepackage{booktabs} 
\usepackage{adjustbox}
\usepackage{multirow}
\usepackage{xcolor}
\usepackage{comment}
\usepackage{color}
\usepackage{xspace}
\usepackage{enumitem}
\usepackage{makecell}
\usepackage{graphicx}
\usepackage{subfig}
\usepackage[utf8]{inputenc}
\usepackage[T1]{fontenc}
\usepackage{tikz-qtree}
\usetikzlibrary{shadows,trees}
\usepackage[
  pass,
]{geometry}
\usepackage{xcolor}
\usepackage{ulem}
\definecolor{uweRGB}{RGB}{210,105,30}

\definecolor{burgundy}{rgb}{0.5, 0.0, 0.13}

\begin{document}
    
\title{Should I Raise the Red Flag? \\
\large An analytic review of anomaly scoring methods toward mitigating false alarms}
\author{\IEEEauthorblockN{Zahra Zohrevand, Uwe Gl\"asser}
\IEEEauthorblockA{School of Computing Science, Simon Fraser University, British Columbia, Canada}
\{zzohreva, glaesser\}@cs.sfu.ca\\[-8pt]
}

\maketitle
\thispagestyle{plain}
\pagestyle{plain}

\begin{abstract}
Nowadays, advanced intrusion detection systems (IDSs) rely on a combination of anomaly detection and signature-based methods. An IDS gathers observations, analyzes behavioral patterns, and reports suspicious events for further investigation. A notorious issue anomaly detection systems (ADSs) and IDSs face is the possibility of high false alarms, which even state-of-the-art systems have not overcome. This is especially a problem with large and complex systems. The number of non-critical alarms can easily overwhelm administrators and increase the likelihood of ignoring future alerts. Mitigation strategies thus aim to avoid raising `too many' false alarms without missing potentially dangerous situations. There are two major categories of false alarm-mitigation strategies: (1) methods that are customized to enhance the quality of anomaly scoring; (2) approaches acting as filtering methods in contexts that aim to decrease false alarm rates. These methods have been widely utilized by many scholars.
Herein, we review and compare the existing techniques for false alarm mitigation in ADSs. We also examine the use of promising techniques in signature-based IDS and other relevant contexts, such as commercial security information and event management tools, which are promising for ADSs. We conclude by highlighting promising directions for future research. 
\end{abstract}


\section{Introduction}
In adversarial settings, cybersecurity systems like intrusion detection systems or insider threat detectors routinely process enormous volumes of heterogeneous log data needed to perform detection and prevention functions.
Tracking wide-ranging characteristics requires effective anomaly detection based on ensemble methods of individual detectors or methods that can handle potential feature interactions in near real-time.

\begin{figure}[h]
\center
\includegraphics[width=1\columnwidth]{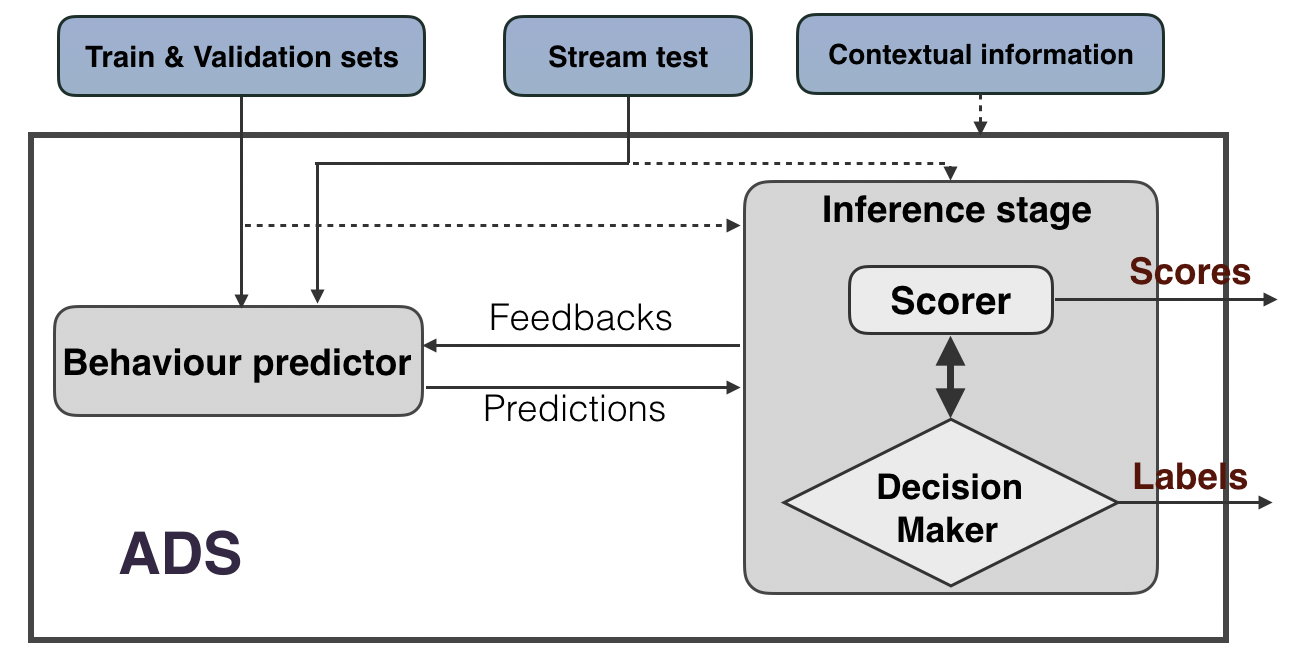}
\caption{Anomaly detection system schema}
\label{fig:ADS_st}
\end{figure}

Learning the normal behavior of complex stochastic systems is a prerequisite for anomaly detection (AD): better estimates of expected observations help detect abnormal cases. 

Advanced deep-learning methods have demonstrated unique capabilities for constructing or forecasting observations.

But anomaly detection reaches far beyond behavior prediction (Figure\ \ref{fig:ADS_st}). A comprehensive anomaly detection system (ADS) must address several challenges: an accurate strategy for scoring and ranking the suspicion level of observations, an optimal threshold of anomaly scores to flag the system status, discard noise or irrelevant outliers, mitigate false alarm rates while maintaining high recall, and explain the causality of anomalous events.
 Advanced behavior prediction methods may decrease uncertainties but do not provide end-to-end anomaly detectors that overcome the challenges to accurately identify suspicious activities. This task calls for extra optimization and pruning, termed the “inference phase” hereafter. 

Uncertainties in anomaly detection---caused by open-ended definitions \cite{hawkins1980identification}, base-rate fallacy \cite{pokrywka2008reducing,sommer2010outside}, ambiguous anomalous boundaries, and noisy environments---make finding optimal thresholds for raising the `red flag' difficult. This holds even for simple scenarios like a one-detector ADS and more so in real-world scenarios with heterogeneous and distributed stream sets and various detectors. A slightly mistuned threshold can cause a huge false alarm rate or routinely miss vicious attacks by labeling them as normal \cite{bridges2017setting}.

Due to subtle differences between {\it anomaly} and {\it malicious event}, high accuracy of anomaly detection does not guarantee reliable detection of malicious activities. While malicious events are generally rare or unusual, simply applying AD to flag them can yield way too many false positives. The real task is puzzling out anomalies of interest; among those, malicious activities are the anomalies caused by adversarial actors aiming to disrupt or physically harm a system \cite{tandon2009tracking}.

Adversaries usually try to blend in with the distribution of normal points \cite{emmott2013systematic}, for instance by intentional  swamping  and  masking  of events. When attacks are not confined to extreme outliers or when the extreme outliers are not anomalous, it is difficult to distinguish anomalies of interest from normal points or inconsequential outliers. Anomalous points are expected to be mostly isolated; in some contexts though, anomalies generated by stealthy adversarial activities may mask or cancel out each other. 
ADSs that focus on the uniqueness of observations to find and score anomalies do not efficiently detect attacks. Also, there is a risk of clustered anomalies, which may occur if most attacks prompt the same type of reaction in the system. Thus, such observations might not appear as isolated anomalies in many rarity-based techniques \cite{chandola2009anomaly}. In particular, if anomalous points are tightly clustered, they would not be detectable by density-based methods.

Additionally, false positives may originate from inaccurate or abrupt spikes in the error rate because of noisy data or mistuned models. For example, a data-driven behavior predictor may miss rare periodic patterns, resulting in sharp spikes in error values even when the behavior is normal \cite{shipmon2017time}.
It is important to review strategies in behavior modeling techniques that decrease the likelihood of generating such anomalies. 

For all mentioned scenarios it is critical to know the related false alarm mitigation approaches. A comprehensive survey on anomaly detection techniques should provide the readers with pertinent information regarding the end-to-end process, including both behavior prediction and inference phases. Given such intuition, the readers would be able to pick and combine the methods based on their strength and the applied context.

\textbf{Related Work.}

Considerable work on anomaly detection originates from statistical research \cite{rousseeuw2005robust, hawkins1980identification, bakar2006comparative}. Studies from computer science have reviewed and surveyed  computational AD concepts \cite{agyemang2006comprehensive, patcha2007overview, hodge2004survey}; the comprehensive research survey in\ \cite{chandola2009anomaly} deeply analyzes pros and cons of traditional AD methods.
Other studies have collected and reviewed state-of-the-art methods in various application contexts. For example, time series AD methods have been thoroughly analyzed and categorized in \ \cite{gupta2014outlier} based on their fundamental strategies and input data types. Other studies have surveyed relevant topics in very specific domains, e.g., \cite{kwon2017survey} provide a comprehensive survey of deep learning based methods for cyber-intrusion detection; a broad review of deep AD for fraud detection is presented in \ \cite{adewumi2017survey} and internet of things (IoT) related AD is reviewed in\ \cite{mohammadi2018deep}. 

\textbf{Motivation and Challenges.} 
Behavior modeling applied in the training of anomaly detection algorithms has caught considerable attention in most review and survey papers, whereas the decision-making process of post pruning, threshold setting, anomaly scoring, and labeling, have widely been neglected.
High false alarm rates tend to confuse data analysts when trying to distinguish normal from anomalous (erroneous) events by just relying on predictive models. This approach is very risky due to uncertain boundaries, continuously evolving behavior, and potential data drifts. And, `chasing ghosts' is a waste of valuable resources after all.

Following a methodical approach, we review false alarm mitigation methods in anomaly detection contexts. The major contributions of our study are:
\begin{itemize}[leftmargin=*]
    \item Building upon the extensive research and surveys on prediction and profiling based AD by significantly expanding the discussion toward anomaly scoring, threshold-setting techniques, and collective analysis, which can contribute to false alarm mitigation. We, 
also, identify unique assumptions regarding the nature of anomalies made by each technique. The combination of such assumptions and pruning steps is critical to discover the failure and successful contexts for that technique. 
    \item 
     Gathering a broad overview of the criteria that anomaly detection and anomaly scoring methods should address to be applicable to real-world problems. Such evaluations help to distinguish these techniques in a more precise realistic way and based on their abilities in finding anomalies which are inherently more complex.
\end{itemize}

\textbf{Organization.} The remainder of this paper is organized as follows. Section~\ref{sec:ProblemDefinition} first explores various anomaly definitions and interpretations, then provides an informal description of  concepts related to anomaly detection on which the rest of the paper relies.
Next, Section~\ref{sec:CCADS} formulates and justifies a collection of requirements which should be addressed in anomaly detection. 
In the next step, Section~\ref{sec:AFS_scaling} analytically and comprehensively reviews the methods to scale the false alarms rate, from the initial scoring stage to the decision making stage for raising the red flag. Section~\ref{sec:post_hoc} continues the false positive mitigation topic in a {\it post hoc} analysis. 
Finally, Section~\ref{sec:ResearchQuestion} highlights research questions and directions for future research.

\section{Basic definitions}
\label{sec:ProblemDefinition}
We start by drilling down into different meanings of anomalies, followed by defining anomaly detection, anomaly scoring and recalling other key concepts and requirements involved in false alarm mitigation for anomaly detection systems.

\subsection{What is an anomaly?}
A general definition of anomaly (outlier) builds on Hawkins' statement in \cite{hawkins1980identification}: "An outlier is an observation which deviates so much from the other observations as to arouse suspicions that it was generated by a different mechanism". The ambiguity in this abstract view leads to divergent goals of AD systems, which can strongly influence the results. Interpretations of anomaly include:

\begin{enumerate}[leftmargin=*]
    \item  \textbf{Anomalies are rare events}. Since unusual cases do not happen frequently, anomalies can be considered rare events. Adopting this interpretation means to find a soft or hard threshold for frequencies. A poor estimation of frequency may cause a huge rate of false positives/negatives. 
 Also, results of methods focused on the rarity score of data points are often not comparable because rarity is hugely dependent on the AD algorithm.

    \item \textbf{Anomalies are distinct events}. Based on this description, any odd event is anomalous. Being different is not meaningful without having some implicit probability that shows the rank of belonging to a distribution or model. So, a threshold is required for determining "to what extent?".
    For example, the clustering methods \cite{portnoy2000intrusion} compute the point's anomaly scores based on their distance to the closest clusters, or the sparsity level of the cluster which it belongs to. Thus, they implicitly consider the distance or density as their differentiating measure.
    
    \item \textbf{Anomalies are abnormal events}. The observations which diverge from normal expectations are anomalous. Same to the previous definitions, finding the divergence degree is ambiguous and might be highly challenging. Also, the normal data should be labelled, while it is not the case in many the target domains like IDS \cite{ferragut2012new}.
\end{enumerate}

But, none of the above interpretations of anomaly detection are acquisitive enough to describe various types of anomalies and suspicious events in different domains.

Based on \cite{ferragut2012new}, an ideal definition of anomaly should be applicable to all of the possible distributions without extra effort. Moreover, it should provide the possibility of comparing the anomalous degree of one target variable versus the anomalous level of the other variables. 
Thus, we provide our definition of an anomaly in a suspicious detection context based on the aforementioned specifications. ``An event is called a {\it suspicious anomaly} if it is highly distinct in terms of feature values, or clustered but unobserved previously, and persistent or close to the previous suspicious cases to reduce their noise likelihood."

\subsection{What is anomaly scoring?}
An ADS aims to order anomalies based on their anomalous score that assigns to the data points. Transferring the original order in feature space through a scoring function $S_{AD} : X \rightarrow R+$, is one of the very basic methods, which assigns smaller scores to the more anomalous points \cite{zuo2000general}. 

\subsection{Data drift and abrupt evolution of data}
The term data drift implies that the statistical properties of the target variable or even input data have changed in the way that the predictive model is decayed and may lose its accuracy level \cite{vzliobaite2010learning}. For instance, customers' behavior in power consumption may change over time because of many reasons like restructuring their internal network; consequently, the consumption predictor is likely to become less and less accurate over time. Generally speaking, it is hard to determine the exact rate of data drifts in an unsupervised anomaly detection context.

\section{Challenges in Anomaly Detection}
\label{sec:CCADS}

Anomaly scoring is challenging for many reasons including high dimensional spaces, stochastic behavior, potential data drift, seasonality or highly irregular data observation rates, uncertain environments, mixed data types, bounded available data, varying lags in the emergence of anomalous behavior in one dimension compared to the others, and unknown hidden factors. Therefore, strong anomaly detection and scoring methods should be capable to address these challenges sufficiently using observed data \cite{veasey2014anomaly}. AD methods are mainly evaluated according to their \textit{detection rate} (i.e., the ratio of correctly detected anomalies to total ones), and \textit{false-alarm rate} (i.e., the ratio of misclassified normal data points to the total number of normal points). However, we have devised and summarized a set of meta-criteria \cite{emmott2015meta, emmott2013systematic} for anomaly scoring as well as corresponding measures to evaluate the strength and performance of the algorithms in addressing the specific circumstances of cyberattack detection and protection.

\subsection{Masking effect}

One anomalous point \textit{masks} a second anomaly if the latter can be considered an anomaly only by itself but not in the presence of the first point. Thus, a \textit{masking effect} may occur if the estimated mean and covariance are skewed toward a cluster of outlying observations such that the outlying point is not sufficiently far from the mean that it can be detected \cite{ben2005outlier}.  

\begin{figure}[h]
\center
\includegraphics[width=0.9\columnwidth]{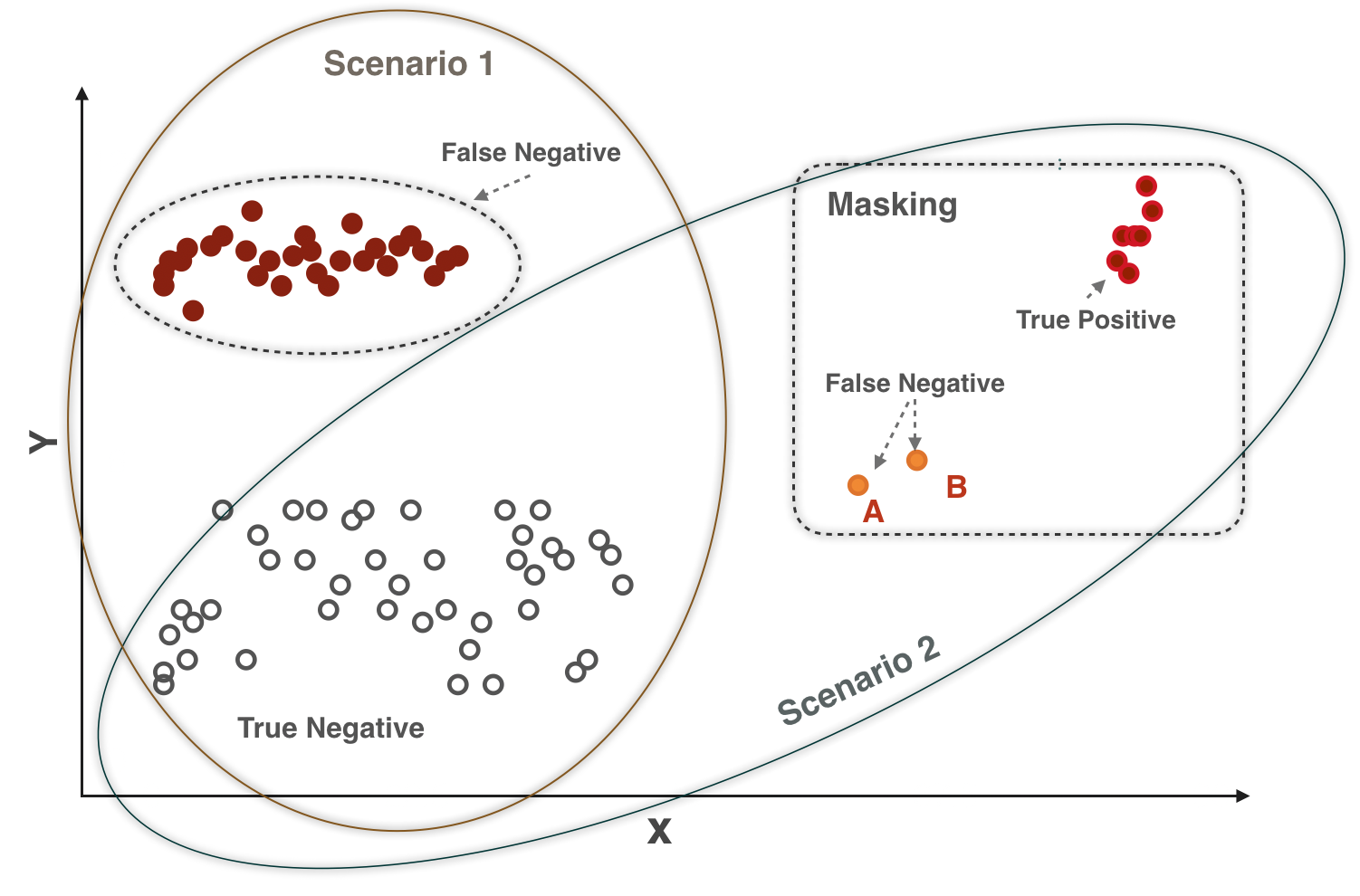}
\caption{An example of rarity-assumption based masking}
\label{fig:rarity_mask}
\vspace{-3.5mm}
\end{figure}

\begin{figure}[h]
\center
\includegraphics[width=0.9\columnwidth]{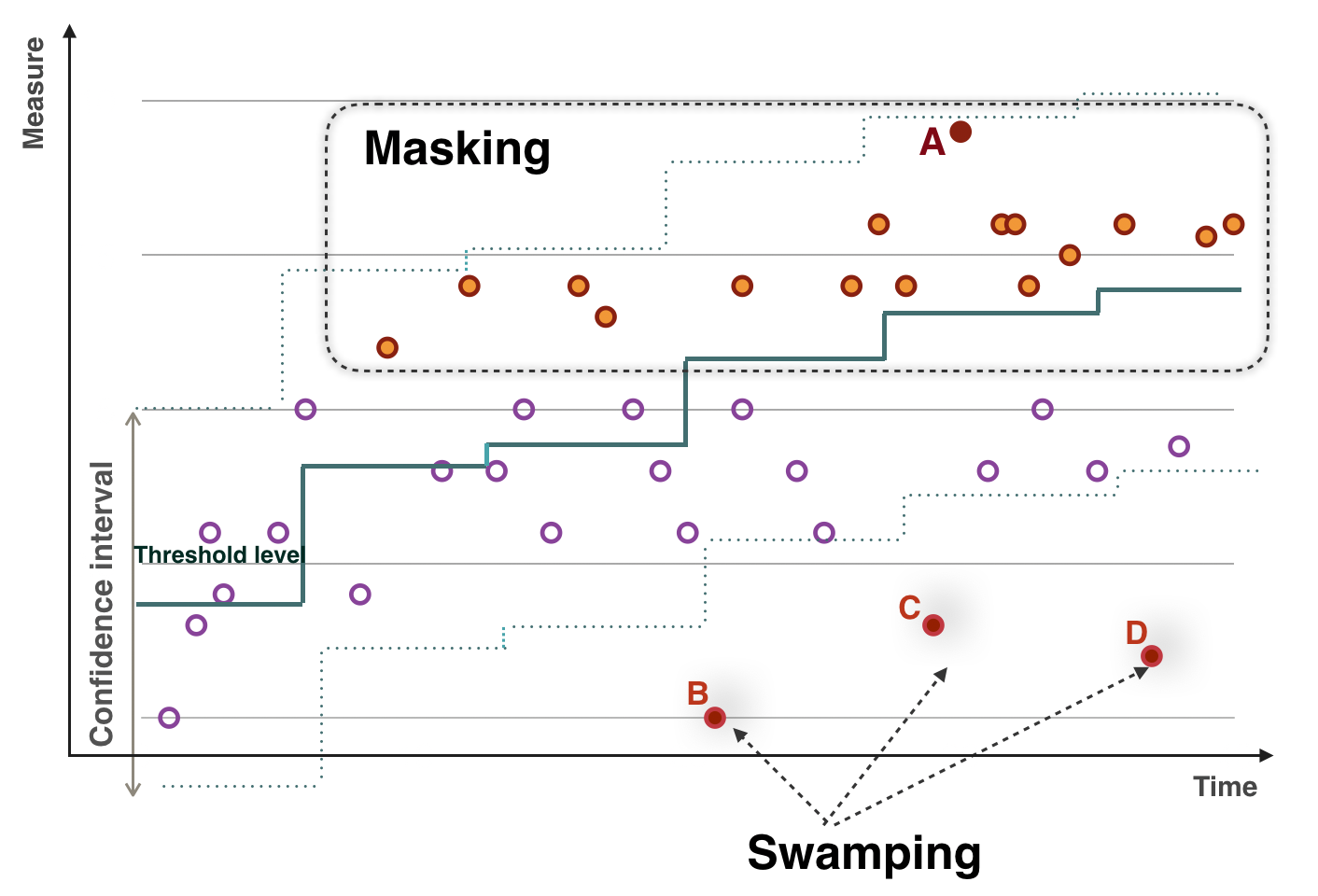}
\caption{A simple example of Masking and Swamping effects}
\label{fig:swamp_mask}
\vspace{-3mm}
\end{figure}

As a toy example of this scenario, a method's behavior that dynamically updates its threshold value is shown in Figure \ref{fig:swamp_mask}. Here, an attacker, aware of this ADS's strategy, gradually poisons the system; i.e. the attacker intentionally feeds the system with fake data in the range of confidence interval but different enough to make the ADS shift its threshold value. Thus, the event (\textit{A}) is in fact a masked attack (\textit{A}) missed by the manipulated ADS. 
Also, if the ADS looks for too few isolated cases, clustered anomalous points can influence the statistics so that none are declared anomalies \cite{liu2008isolation}. As an example, the rarity-based ADS, shown in Figure \ref{fig:rarity_mask}, has missed \textit{A} and \textit{B} because of the limited number of anomalous points assumption. 
\textit{semantic variation}, which represents the degree to which anomalies are dissimilar \cite{lavin2015evaluating}, is one the measures for evaluating the ability of an ADS to handle masking effects.

\subsection{Swamping effect}
\textit{Swamping effect}, the reverse of the masking effect, occurs if the swamped events can be considered an anomaly only in the presence of other events. 
For instance, the false-positive events \textit{B–D} in Figure \ref{fig:swamp_mask}, are swamped by the orange observations. 

Outlying groups which skew the mean and covariance estimates toward themselves can push away normal events from the shifted mean to be isolated as anomalies \cite{ben2005outlier}.
If an ADS overestimates the number of anomalies in a dataset, it can be influenced by the swamping effect. 

\textit{Point difficulty} is the measure which evaluates the swamping effect \cite{emmott2013systematic}. The point difficulty of each observation is measurable based on its likelihood of belonging to the other class in comparison to the current class label. Ideally, a method should be able to detect anomalies with higher point difficulty rates.

\subsection{Variable frequency of anomalies}
A rarity-based ADS normally performs well if the frequency of anomalies is low, ranging from 1--10\%, but may fail in other scenarios, e.g.\ DoS attacks which are more frequent (>30\%) \cite{kim2012robust, liu2008isolation}. \textit{Scenario 1} in Figure \ref{fig:rarity_mask} illustrates an ADS which misses a group of anomalies only because of the rarity assumption.
The reliability of ADSs under variable frequency conditions is measurable based on its tolerance level under different degrees of \textit{relative frequency} without losing accuracy. The measure relative frequency (contamination rate) is defined based on the proportion of anomalous data instances  \cite{emmott2013systematic}.

\subsection{Curse of dimensionality}
Access to more features and detectors decreases the risk of missing influential information when performing an ML task, but it is associated with fundamental research problems \cite{zimek2012survey}, like \textit{Irrelevant features}, \textit{Concentration of scores and distances},  \textit{Incomparable and uninterpretable scores}, and \textit{Exponential search space}.
For example, each irrelevant feature increases the space dimensionality; and the sample size required by (naïve) density estimation methods tends to scale exponentially with dimensionality. 
Irrelevant features decrease precision and increase false alarm rates. With increasing dimensionality, the data domain space grows. So, the normal points may be pulled away from the others and be trapped in unimportant tails \cite{emmott2015meta}, while the anomalous cases are covered under the same unimportant similarities.
\textit{Feature ranking} is a practical measure to verify a model's reliability for ignoring irrelevant features by paying attention to features according to their importance \cite{das2018a, amarasinghe2018toward}. This evaluation metric is only applicable to supervised data and highly dependent on the applied method.

\subsection{Lag of emergence}
In complex systems with complicated feature associations and causal relations the source of anomaly may trigger system features in various ways. For instance, some features may react to events sooner than others; therefore, the anomaly flag will be raised several times with various lags.
Causal relations are one of the main reasons behind this fact, which generally leads to scenarios that dependent features indicate the same events with variable delays. Also, having irregular sampling rates in different dimensions is another potential reason for such lags. Features with a lower sampling rate (longer interval) may indicate extreme or anomalous events which have been already deciphered from other high-frequency features \cite{veasey2014anomaly}.
Thus, an ADS is expected to aggregate the results obtained from heterogeneous subsystems and features to discover the exact point of a suspicious anomaly, rather than overwhelming the users with frequent alarms over the course of time. 

\subsection{Domain specific criteria}
Defining domain-specific measures may help evaluate ADS capabilities in very specific target domains. For example, \cite{lazarevic2003comparative} applies "{\it burst detection rate (BDR)}" to capture potential bursts which indicate attacks involving many network connections. This measure represents the ratio of the total number of intrusive network connections with a higher score than the threshold to the total number of intrusive network connections within attack intervals. 

\section{Automatic false alarm mitigation}
\label{sec:AFS_scaling}

Stochastic and evolving normal behavior complicates the AD process. On the one hand, behavior predictors may not be able to find the exact underlying patterns of the data. On the other hand, a small mistake in the scoring or ranking process can lead to a huge amount of false positives or false negatives.

This section reviews various approaches taken by statistical, data mining, or ML methods to decrease false alarm rates by scaling anomaly scores. Some of those perform scoring and ranking simultaneously as a unified task, whereas others assign a set of initial scores to observations, then rank them based on other potential available sources of information or collective analysis.
We expect the anomaly scores assigned by an ADS to be at least in a {\it weak ordering} \cite{roberts2009},

so events can be ranked based on their deviation from the expectation.

\subsection{Improved individual scoring}
This category includes methods and techniques contributing to a better scoring than simple anomaly scoring based on the error vector, i.e.\ the differences between the observations and the expectations generated by a behavior predictor.
\smallskip
\subsubsection{\textbf{\textit{Probability-based scoring}}}
Assuming a particular distribution for the dataset, the anomaly score of the observations is computable based on their probabilities and data statistics. 
Intuitively, if $x$ is less likely to happen, it is more anomalous. In other words, an anomaly score $A(x)$ respects the distribution  ($f$), then $A(y) \leq A(x) \iff f(x) \leq f(y)$.

This approach is deemed reasonable in under control conditions, but it is not a powerful technique in real-world scenarios \cite{bridges2017setting, ferragut2012new} because: 1) Data dispersion is full of noise and often far from the known statistical distributions; 2) It may ignore the rarity condition, i.e.\ with the same threshold, long tail distributions provoke many red flags versus shorter tails; also, 3) Obtained scores are not comparable and agreeable in complex configurations with multiple cooperative detectors.
\noindent Variations of probability-based scoring include:

\smallskip
\textit{1.1) Bits of rarity.}
Let's assume a model that provides a probability density distribution of values or errors \cite{tandon2009tracking}. A bits of rarity based anomaly score of an event $x$, with the probability density or mass function $f$, can be defined as:  
$$R_f(x) = -log_2(P_f(x))$$
This technique uses a one-to-one transformation of the predictor results to present a more explainable ranking. The log-scale transformation of scores helps to stabilize the computations and distribute the original probabilities, $0 \leq P_f(x) \leq 1$, to a larger range of values. Also, the negative sign assigns a higher anomaly score to more diverged events. Nonetheless, expanding the scores in the range of natural numbers however leads to a huge difference in the computed anomaly scores by various detectors, so that their results may not be comparable. 
 
\smallskip
\textit{1.2) P-value scoring.} P-value of statistical tests is another traditional technique for determining potential outliers \cite{schervish1996p}. Because of its independence from probabilistic distribution functions (PDF), using the p-value to find anomalies is intuitive. Also, the p-value can rank all the observations based on their dissimilarity and rarity, as two major interpretation of anomalies; it puts a sharp bound on the {\it alert rate} only based on the probabilistic description, so, it narrows down the frequency of alarms for any random distribution. 
But the p-value only focuses on extreme cases as a subset of all target anomalies; i.e. the ones that lie outside the convex hull of most of the distribution mass \cite{veasey2014anomaly, bridges2017setting}. Also, it assumes that no other alternative hypothesis is available, which is often not the case in real-world settings. After all, the choice of the significance level to reject the null-hypothesis is crucial.

\smallskip
\textit{1.3) Bits of meta-rarity.}
By not only considering an event's rarity but also the infrequency of its rarity level, bits of meta-rarity makes anomaly scores directly comparable \cite{ferragut2012new}. The formal definition of this measure is as follows:
$$ A_f(x) = -log_2(P_f(f(X)\leq f(x)))$$
Thus, it provides a strict weak ordering of the observations based on their abnormality level, so that $ x >_a y$, if and only if $A_f(x) > A_f(y)$.
Also, the assigned anomaly degree would exceed the given threshold value of $\alpha$ with a chance of not greater than $2^{-\alpha}$ 
($P_f(A_f(x)>\alpha) \leq 2^{-\alpha}$).
Due to being dependent only on $\alpha$, rather than also on the data distribution ($f$), the number of false alarms generated by this approach can be regulated.
Moreover, considering the same theory, this approach allows for comparing $X$ and $Y$, generated by $f$ and $g$, based on $A_f(X)$ and $A_g(X)$, respectively.

\begin{figure}[h]
\center
\includegraphics[width=0.7\columnwidth]{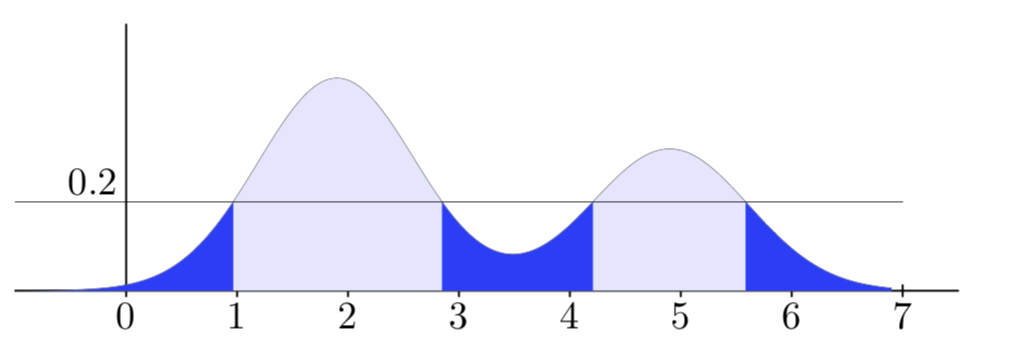}
\caption{Bits of meta-rarity of 0.2 in a mixture of normal distribution \cite{veasey2014anomaly}}
\label{fig:mix_gaus}
\vspace{-3.5mm}
\end{figure}

As Figure \ref{fig:mix_gaus} illustrates, this technique is capable of finding anomalous areas which are not necessarily extreme values like the very low probability points in a Gaussian mixture model. 
But it suffers from two main drawbacks: 1) The value of $\alpha$ can hugely affect the false positive rates; 2) Bits of meta-rarity generally ignores clustered anomalies.

\smallskip
\subsubsection{\textbf{\textit{Q-function based scoring}}}

By presuming a Gaussian distribution for the error vector obtained from predictions and normal observations, some studies fit the anomaly scores based on a Q-function as a tail distribution function \cite{goldstein2016comparative, shipmon2017time, ahmad2017unsupervised}. 
For example, a Q-function scoring is used in \cite{malhotra2015long} to analyze error values. %
At first, a stacked LSTM model is trained to predict the next $l$ values for $d$ dimensions of input variables. Then, as the predictor slides through the observations at $\{t-l, ..., t-1\}$, it forecasts the values of each time point, $x(t)$ for $l$ times. Next, the ADS generates an error vector, $e(t)$, as the difference between $x(t)$ and its predicted value at different time points ($t-j$) as:

($e(t) = [e(t)_{11}, ..e(t)_{1l}, .., e(t)_{d1},..., e(t)_{dl}]$).

Then, it fits a multivariate Gaussian distribution to the error vectors ($N = N(\mu, \sigma )$). 
This means the likelihood ($p(t)$) of observing an error vector ($e(t)$) is equal to the value of $N$ at $t$.

Thus, an observation $x(t)$ is \textit{anomalous} if $p(t) < \tau$; $\tau$ is computed by maximizing the $F\beta-score$ based on an unobserved validation set.

A fast comparison between new error values and the compact representations of the prior cases is one of the main advantages of this method \cite{shipmon2017time, ahmad2017unsupervised}. 
However, the normal error distribution will be violated if the error values would not be random, which is likely in many data-driven methods \cite{hundman2018detecting}.
\smallskip
\subsubsection{\textbf{\textit{Similarity based scoring}}}
These types of methods compute the anomaly score for new observations based on their distance to other groups of observations like the set of their $k$ neighbors.
Distance-based techniques confront the threshold setting problem in the early stages of the AD process by looking for the appropriate distance (similarity) to distinguish the far points from the close ones. Thus, this step is comparable to behavior modeling in profile-based ADSs. 
Distance metrics can be grouped into the following three categories \cite{weller2015survey}:
\begin{itemize}[leftmargin=*]
    \item \textit{Power distances.} Distance measures which use a formula mathematically equivalent to the power of $(p, r)$ as: 
    $$ Distance(X, Y) = \left (  \sum_{i=1}^{n}\left | x_i - y_i \right |^p\right )^{\frac {1}{r}}$$
    For example, Manhattan and weighted Euclidean distance should be included in this category. Advanced power distances are more practical but not necessarily intuitive and compatible with the physical distance concept \cite{deza2009encyclopedia}.
    \item \textit{Distances on distribution laws.} These describe the distance measures based on the probability distribution of the dataset, like the Bhattacharya coefficient \cite{patra2015new} or $\chi^2$ distance \cite{deza2009encyclopedia},
    \item \textit{Correlation similarities.} This group characterizes the correlation between two datasets as a measure of similarity or distance, such as Kendall's $\tau$ rank correlation and learning vector quantization \cite{kohonen1990self}.
\end{itemize}

Utilizing appropriate distance metrics corresponding to the data distribution can improve the anomaly scoring phase. For example, due to the difference in feature distributions, Euclidean distance is not the right metric to capture the real distance of points from the mean of normal data. Thus, some anomaly detection studies \cite{wang2004anomalous, lazarevic2003comparative} take advantage of Mahalanobis metric \cite{mccrae1987creativity}, which is able to take into account the variable's variance and covariance besides the average value. 
Scoring anomalies using the similarity evaluation measures from the second and third groups is almost comparable to the probability-based and information theory-based scoring techniques, respectively, which are discussed in their related sections. In sum, distribution-aware measures are valuable regarding their improvements in the similarity evaluation phase \cite{weller2015survey} that contributes to a more reliable estimation of the divergence of new observations from normal expectations.

\smallskip
\subsubsection{\textbf{\textit{Extreme value theorem}}}
\label{sec:EVT}
The extreme value theorem is explored in \cite{siffer2017anomaly} to find distribution-independent bounds on the rate of extreme values for univariate numerical time series. This technique does not require any manual threshold setting but needs one parameter as the {\it risk factor} to control the number of false positives. 
The law of extreme values states that extreme events have similar kinds of distributions, regardless of the main data distribution as long as it is standard \cite{fisher1928limiting}:
$$G_\gamma = x \rightarrow exp(-(1+\gamma x)^{\frac{-1}{\gamma}}), \gamma \in R , 1+\gamma x>0$$
$\gamma$ is called {\it extreme value index} and depends on the original distribution; e.g. it is zero for Gaussian ($N(0, 1)$). When events are extreme, the shape of distribution tails are almost similar, so $G_\gamma$ can represent all of them. 

Based on this theory, a streaming anomaly detector, called SPOT, is proposed in  \cite{siffer2017anomaly}. 
In the first step, SPOT computes $z_q$ as the hard and $t$ as the soft threshold by fitting a generalized Pareto distribution and then utilizing an appropriate extreme value distribution.

\begin{figure}[h]
\center
\includegraphics[width=0.9\columnwidth]{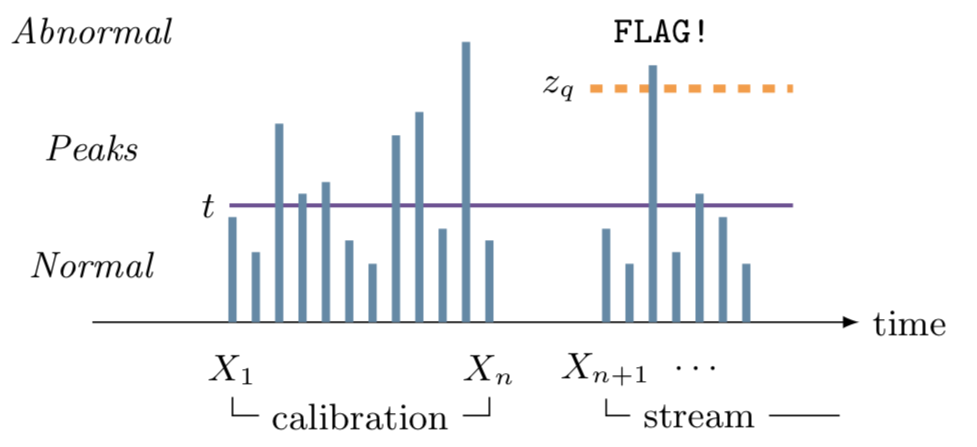}
\caption{Updating anomaly scores in stationary streams \cite{siffer2017anomaly}}
\label{fig:st_update}
\vspace{-2mm}
\end{figure}

As Figure \ref{fig:st_update} illustrates, SPOT flags the data points that exceed the $z_q$ threshold as abnormal, while it keeps updating $z_q$ based on the rest of the observations (non-abnormal cases). Each non-abnormal case might fit one of the following scenarios: 1) \textit{Peak case.} It is greater than the initial threshold $t$, so adds the excess to the set of peaks and updates the value of $z_q$, 2) \textit{Normal case.} It is a common value.

\subsection{Unified scoring}
The issue of non-comparability and non-interpretability of different ADS results is targeted in \cite{kriegel2011interpreting}. 
Unified scoring \cite{kriegel2011interpreting} converts any arbitrary "anomaly factor" to the interpretable range of $[0, 1]$ as an indicator of the abnormality probability.
Defined based on Hawkins' idea, this unification transformation method includes two steps, where either step might be optional (depending on the type of score ($S$)): 1) Regularization, basically maps a score $S$ to the interval $[0, \infty)$, so that ${Reg}_{S(o)} \approx 0$ represents inliers and ${Reg}_{S(o)} \gg 0$ indicates outliers; 2) A normalization to transform a score into the interval $[0,1]$.
The applied transformation method should be {\it ranking-stable}, which means it should not change the ordering of the original scores.\\
However, the authors do not propose any solid algorithm to apply the mentioned mapping to arbitrary AD scoring and they only offer some abstract generic hints.

\subsection{M-estimation scoring}
\label{sec:M-estim}
Based on the fact that the means of tail estimations indicate the anomalous level of data points in univariate domain spaces, M-estimator \cite{clemenccon2013scoring, clemenccon2018mass} is proposed to simulate the same property in higher dimensional space. This technique can address unsupervised scoring and ranking of anomalies in multivariate domain spaces.
M-estimator captures the extreme behavior of the high-dimensional random vector $X$ based on the univariate variable $s(X)$, which can be summarized by its tail behavior near zero such that the smaller the score $s(x)$, the more abnormal is the observation $x$.

\begin{figure}[h]
\center
\includegraphics[width=0.9\columnwidth]{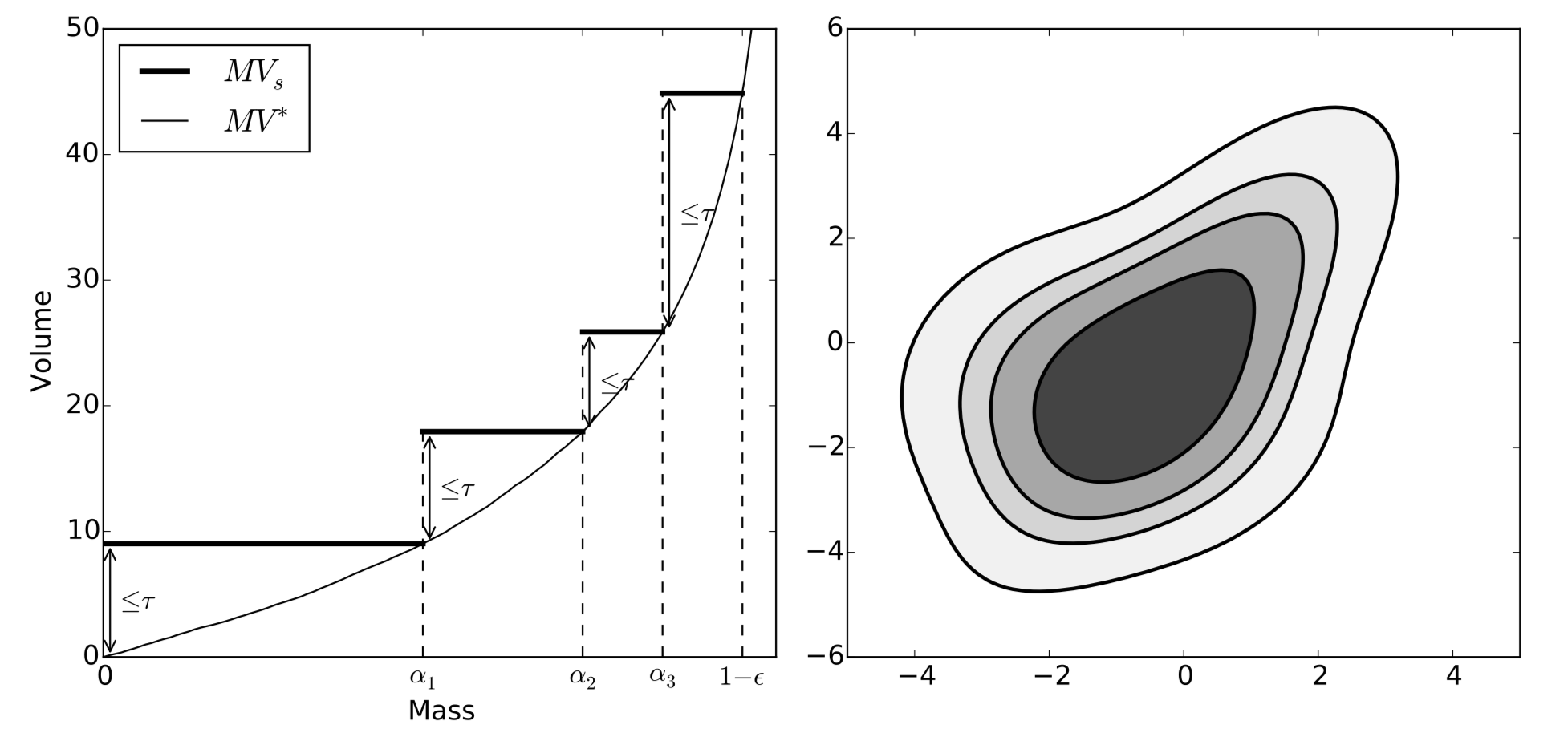}
\caption{left: Piece-wise adaptive approximation of $MV^{\ast}$ and right: associated piece-wise scoring function \cite{clemenccon2018mass}}
\label{fig:pw-MV-opt}
\vspace{-2mm}
\end{figure}

This technique uses the mass volume ($MV$) curve as a functional performance criterion to estimate the density function. Then, it provides a strategy to build a scoring function $s^{(x)}$, where its $MV$ curve is asymptotically close to the empirical estimate of the optimum mass volume (${MV}^{\ast}$). In the next step, the functional criterion is optimized based on a set of piecewise constant scoring functions. 
In the end, the feature space is overlaid with a few well-chosen empirical minimum volume sets as Figure \ref{fig:pw-MV-opt} illustrates.

\subsection{Improved Threshold Computation}
This section reviews a set of customized threshold setting strategies that can help mitigating false alarm rates.
\smallskip
\subsubsection{\textbf{Receiver Operating Characteristic curve (ROC)}}
This is a graph showing the performance of a classification model at all potential thresholds. ROC or Precision-Recall Curve are some of the common techniques used by supervised AD methods to find the best error threshold for discretizing the ranked observations. Since this technique is not suitable for an unsupervised context, it should be replaced with the MV curve \cite{clemenccon2013scoring, clemenccon2018mass, goix2016evaluate}, as explained in \ref{sec:M-estim}.

\smallskip
\subsubsection{\textbf{Dynamic threshold}}
In \cite{hundman2018detecting}, a dynamic threshold based approach is proposed for evaluating residuals to address non-stationarity and noise issues in data streams. It includes following two main steps:\\
\noindent{\textit{a. Error computation and smoothing.}} It computes a one-dimensional error vector based on the expectation and observation values. Then, it smooths this error vector using exponentially-weighted average to get  $error_s$ :

$${error}_s =[e_s(t-h),...,e_s(t-1), e_s(t)]$$ 
where $h$ determines the number of historical error values used to evaluate the current errors. 
\noindent{\textit{b. Threshold calculation and anomaly scoring.}} It optimizes a threshold, $\epsilon$, so that removing all the values above it lead to the greatest percent decrease in the mean and standard deviation of the smoothed errors $e_s$. Then, the normalized score of the highest smoothed error, $e_s(i)$,  in each sequence of anomalous errors will be determined based on its distance from the chosen threshold. 

However, finding the best value for $z$ is extremely context dependent. So, the problem of threshold optimization remains, but in smaller scales. Also, adding gradual anomalies to data can lead the system to increase the threshold value and miss the real attacks.

\subsection{Sequence based scoring}
Having a bird's-eye view from the whole sequence can lead to a higher recall, but lower false alarm rates \cite{ahmed2017thwarting, zohrevand2016hidden,zohrevand2020dynamic}. However, it comes in the cost of losing real-time response, unless the ADS gradually performs anomaly detection \cite{zohrevand2016hidden}. This section reviews three different techniques to apply the collective relation of data points in scoring the whole sequence. 
\smallskip
\subsubsection{\textbf{Information theory based AD}}
Traditionally, information-theoretic measures like {\it (Conditional) Entropy, Relative Entropy}, and {\it Information Gain} were very popular in tracking the anomaly likelihood in the data. For example, the conditional entropy, $H(X|Y)$, of the system call subsequences, can help to determine the suspicious traces \cite{lee2001information}. Also, the computed relative entropy could help to validate the model's quality on the new observations and detect data drifts.
\smallskip
\subsubsection{\textbf{Likelihood ratio method}}
The probability of any data point in this family of strategies is computable considering the precedent observations. Therefore, a sequential anomaly score is a function of the data points' likelihood in the sequence. So, a sequence with a very low generation probability should be marked as an anomaly. Many ADS studies and applications in various areas like intrusion detection and speech recognition have applied different modifications of this strategy. Three high-level methods in this category are: 
\begin{itemize}[leftmargin=*]
\item \textit{Finite State Automata (FSA).} It trains an FSA and if tracing a sequence ($x$) in the FSA ends up to a state without an outgoing edge to the next value in the test sequence, it will be labeled as an anomaly \cite{chandola2009anomaly}.
\item \textit{Markov Models.} It obtains the conditional probability of the observed symbols and their transition to each other. Anomalous series will be distinguished based on their lower generation probability \cite{sun2006mining}.
\item  \textit{Hidden Markov models (HMM).} An HMM learns the underlying patterns of training sequences. Then, the likelihood of a test sequence, generated by the HMM, will be verified using decoding algorithms like {\it Viterbi algorithm} \cite{zohrevand2016hidden}.
\end{itemize}

A distinguishing property of this group of techniques is their ability to consider observations' correlation and order to score the whole sequence. However, this category of analysis decreases the possibility of detecting short-term abrupt differences as an anomaly.

\subsection{Collective analysis}
The methods described in this section take advantage of some extra information like previous observations, contextual, or correlation information to improve the score or verify the labels assigned to these observations.
\smallskip
\subsubsection{\textbf{Voting-based methods}}
Applying a hybrid of various detectors to detect anomalies is very promising and decreases the chance of raising a false alarm. 
Even, some studies apply voting based on a combination of the current generated alarms and the obtained historical feedback from system administrator to rank alerts. 
For example, a voting technique based on HMM models is proposed in \cite{zohrevand2016hidden} to perform anomaly scoring.
This ADS confirms the anomaly detection results by applying a group of reference windows (RW), in which their context is very similar to the current test window (TW). If both windows have a similar overall transition, the detected anomaly in the lower level is unreasonable, and its anomaly score should be decreased based on the similarity ratio. Otherwise, the anomaly score obtained in leaf nodes is increased according to the inverse of the similarity ratio.
\smallskip
\subsubsection{\textbf{Rolling feature based method}}
 An online collective analysis technique is proposed in \cite{zohrevand2020dynamic} to handle \textit{flagging} of suspicious events based on anomaly detection techniques. It assumes that attacks aiming at severely disrupting the system usually cause lasting cascading effects, so a persistent anomalous interval is more suspicious of being attack than a single strike caused by sensor noise or predictor faults. Thus, this study performs dynamic flagging based on deviation and persistency trade-off. To this end, it computes the moving average ($\bar{M^s}$) of the standardized error vectors ($\bar{Z^s}$) within the temporal window ($w_{p}$), selected as the average number of steps that an anomalous event is typically expected to last to be associated with an attack.
$${m^s_t = m^s_{t-1} + (z^s_t - z^s_{t-w_p})/{w_p}} \vspace{-1mm}$$
Thus, the first observation ($x^s_i$) with $m^s_t$ value higher than the threshold triggers raising a red flag for a local attack, provided that at least 50\% of points have been detected as anomalous in the considered $w_{p}$, which is the direct predecessor of $x^s_i$. 

\smallskip
\subsubsection{\textbf{Scoring in different levels of granularity}}
Anomaly scoring in different levels of granularity allows traceability down to the finer granularity level and decreases the chance of missing those low-level patterns. The obtained anomalies in lower levels can be grouped to ultimately find anomalous cases in the subsystem level. A very logical break-down in the network IDSs is modeling detectors in the node and network level to be able to trace both focused and distributed attacks.
Another approach, in this context, is applying break-down and aggregation on the time dimension to trace and balance the scores based on their short and long-term stability. For instance, \cite{zohrevand2016hidden} applies a hierarchical confirmation procedure to improve accuracy. From another point of view, this method applies global Markovian models in higher levels of the hierarchy to verify the leaf nodes' results. The overall approach in the higher levels is very similar to the lowest level, except that the aim of these comparisons focuses on the state transitions instead of the real values of observations. 
The hierarchy of granularity can be defined based on various features, like the space dimension. For example, to find the Spatio-temporal objects whose thematic attributes are significantly different from those of the other objects, \cite{cheng2006multiscale} proposes a method based on multi-granularity and cluster differentiation.
\smallskip
\subsubsection{\textbf{Alarm correlation}}
Constructing the potential attack scenarios based on the data aggregation is one of the other strategies in mitigating false alarms.
The aggregation process can be performed by grouping a bunch of alarms possibly generated by different detectors or in different places of network or sequence and then reconstructing the attack scenarios. 
Performing correlation analysis and generating the possible scenario help to extract some concrete and interpretable inferences to decrease the false-alarm rates. Correlation analysis can be considered in different levels of abstraction, like the correlation of events from the same or heterogeneous detectors, correlation of events in one detector through time dimension, correlation of events through different nodes in network \cite{hubballi2014false}. Some alarm correlation analysis techniques are:

\begin{itemize}[leftmargin=*]
\item \textit{Multi-step correlation.} By assuming that a sequence of actions is required to conduct a malicious mission in the system, finding a correlation of the observed anomalies may help to assert the malicious event, before it happens \cite{ning2002analyzing, hubballi2014false}. One of the relevant techniques is applying frequent pattern mining to track frequent IDS alarm combinations as the indicators of malicious sequences \cite{sadoddin2009incremental}.
\item \textit{Causal relation based correlation}. It verifies the causality correlation between existing variables and detectors. For instance, by considering the Bayesian network of nodes as alarms and edges as relationships obtained from the time-based coincidence of alerts and their mutual information, the system can generate hyper-alerts \cite{qin2007discovering,zohrevand2020dynamic}.
\item \textit{Subsystem graph based correlation.} A system typically includes many subsystems with cascading influence on each other. So, an attacker may use a weakly protected subsystem, because of being known as low impact vulnerability, as a footrest to reach the most critical subsystems and servers in the network. Thus, some IDSs identify the possible penetration to critical systems by focusing on the existing dependency and interconnections in the network of systems and balance anomaly scores based on the penetration paths which the events might cause \cite{roschke2011new,valdes2000approach}.
\end{itemize}

\smallskip
\subsubsection{\textbf{Alarm verification}}
The whole idea of such post-analysis approaches is verifying whether the detected unusual events will impact the system. Such verifications help ADS to categorize the detected cases based on their impact seriousness \cite{hubballi2014false,bolzoni2007atlantides}. There are two types of verification mechanisms: \textit{Passive verification} and \textit{Active verification}. The former performs the verification process versus a database including possible success cases, while the latter verifies the generated alarms, in an online manner, which seems more promising to detect zero-day attacks and be applicable in the context of stream data anomaly detection.
However, these strategies may fail if attackers generate some spurious patterns of responses to misguide the IDS to believe that the attacks will fail. For example, in signature-based methods, there is a class of {\it Mimicry attacks} that the attacker sends a fake normal response on behalf of the server. Thus, IDS fails to detect malicious behavior and ignores the alarm \cite{todd2007alert}.

\smallskip
\subsubsection{\textbf{Apply contextual information}}
There are three major ways that utilizing contextual information can help ADS to improve their precision levels:
\begin{itemize}[leftmargin=*]
    \item \textit{Multivariate analysis.} It considers contextual information as the extra features to the existing data and train a multivariate model \cite{zohrevand2016hidden}.
    \item \textit{Error balancing.} It performs an early prediction and readjust the results by benefiting from contextual information \cite{zohrevand2017deep}.
    \item \textit{Post verification.} It utilizes contextual information in the post-verification phase to prune unrelated anomalies and false alarms \cite{radon2015contextual}.
  \end{itemize}

\section{Post-hoc mitigation of false alarms}
\label{sec:post_hoc}
The precision of ML methods is highly dependent on the comprehensiveness of the available observations during the training phase. This issue is even more common in profile-based anomaly detection approaches because they are influenced by training data in two ways: 1) to fit an accurate prediction model, and 2) to set a precise isolation boundary.
This is while, data streams are prone to various drifts (like trend and abrupt evolution), which make the model built on old data inconsistent with the new data and might increase false-positive rates \cite{hundman2018detecting}. 
Thus, this section reviews a set of strategies mainly focused on utilizing user feedback or the history of observations to readjust the threshold for relabeling the data points. These types of techniques are particularly beneficial in non-real-time configurations like fraud detection or other contexts, in which an ADS should prioritize the suspicious cases and there is no need for instant reactions. But some ADSs also take advantage of the obtained knowledge from this step to improve their future scoring process.
  
\subsection{Maximum error value based pruning}
The ADS proposed in \cite{hundman2018detecting} prunes the detected anomalies based on the maximum value of all the observed anomalous points. To this end, this method keeps track of a set, called $e_{max}$, containing top anomalous values ($e_a$) of error sequences, $E_{seq}$, sorted in descending order, plus the maximum smoothed error value that is not anomalous: 
$$e_{max} = max({e_s \in E_{seq} |e_s \in e_a })$$
Then, it goes through this sequence and computes the values of the percentage decrease as:
$$d^{(i)} = (e_{max}^{(i-1)} - e_{max}^{(i)} ) / e_{max}^{(i-1)}  |  i \in {1, 2, ..., (|E_{seq}|+1)}$$
Next, it considers the threshold of $p$ as the minimum percentage decrease to hold the anomalous status.
If at a specific step ($i$) the threshold $p$ is exceeded by $d^{(i)}$, the anomalous status of all the cases before that ($e^{(j)} \in e_{max}  |  j < i)$  remains valid. While, if the $p$ threshold is not met by $d^{(i)}$ and all its subsequent errors ($d^{(i)}, d^{(i+1)}, . . . , d^{(i+|E_{seq} |+1)}$), those smoothed error sequences will be reclassified as normal.

\subsection{Rarity based post pruning}
The main idea in this technique is based on the rarity assumption in the AD context. For example, the method applied in \cite{hundman2018detecting} is configured to consider a minimum frequency for the number of observed anomalies in the same magnitude such that it classifies the future occurrences of this frequent category as normal. The prior anomaly scores for a stream data can be used to set an appropriate threshold, depending on the desired balance between precision and recall. However, this pruning step is highly prone to be misled by the attackers and gives the green card to malicious behaviors.

\subsection{Active-learning based scoring}

If the ADS has a mechanism by which human analysts can label a subset of data, the system can take advantage of the provided labels to set a threshold, $s_{min}$, for a given stream based on the lower and upper bound score of the confirmed anomalies.
ADS proposed in \cite{das2018a} finds anomalies by designing a hyper-plane that passes through the uncertainty regions based on the learned decision boundaries through active learning. The best threshold should be found to customize this hyper-plane.
To minimize learning interaction with the end-user, it assumes that the analyst can only label a limited batch of $b$ instances in each feedback iteration, so this set should be diverse and impact.
This technique is performed in three following steps: 1) Select $Z$  candidates as the top-ranked instances, 2) Selects S compact subspaces that contain $Z$, 3) Selects the set $Q$ ($Q \subset Z$) including $b$ instances, which belong to minimal overlapping regions.

\section{Improved predictor models}
Applying a predictor model \cite{gamboa2017deep, malhotra2015long}, capable of capturing the latent complex patterns of the data, will unquestionably contribute to higher recall and lower false alarm rates. This section briefly outlines several predictors suitable for anomaly detection.

\subsubsection{Robust anomaly detection}
The number of data points in the anomalous clusters or their distances to the normal cases should not influence the decision boundary, otherwise, the ADS may overlook part of the anomalies (masking) or mislabel some normal cases as anomalous (swamping). Thus, some studies \cite{xiong2011direct, DBLP:journals/widm/RousseeuwH11} are focused on improving the quality level of the underlying applied method, like applying robust PCA instead of PCA, or robust matrix factorization, which can end up in better false alarm and recall rates.

\subsubsection{Advanced data driven models}

Traditional algorithms are not scalable enough to handle complex behavioral patterns in big data. Besides, reducing dimensions or selecting the most important features out of thousands of dimensions is not a trivial task for traditional models.
While deep models are qualified to address all of the mentioned challenges. There are a variety of studies which apply different deep models as predictor model to perform anomaly detection: 1) Discriminative models: RNNs like long short term memory (LSTM), gated recurrent units (GRU) \cite{malhotra2015long, nanduri2016anomaly}, convolutional neural networks (CNN) \cite{vinayakumar2017applying}, deep neural networks (DNN) \cite{akhter2012detecting}; 2) Generative models: different versions of auto-encoders \cite{sakurada2014anomaly, an2015variational} and sum-product networks \cite{poon2011sum}; and 3) Generative adversarial networks (GAN) \cite{schlegl2017unsupervised}.
For more detail, we refer the reader to the following research surveys \cite{kwon2017survey, adewumi2017survey, mohammadi2018deep}.

\subsubsection{Hybrid models}
Due to achieving locally optimal results and considering a limited number of hypotheses, none of the existing individual models can be considered a perfect approach. To address this problem, ensemble and hybrid methods that find the optimum non-homogeneous decision boundaries between normal and anomalies \cite{kazienko2013hybrid,das2018a} can be utilized. For example, a hybrid of model-driven and data-driven methods is utilized in many studies to cover the potential weaknesses of each side \cite{khashei2011novel, egrioglu2013fuzzy, zohrevand2017deep}. Also, deep hybrid models \cite{erfani2016high,javaid2016deep,weston2012deep,poultney2007efficient} use an unsupervised technique prior to the task of interest to learn the reduced representative features.

\section{research questions}
\label{sec:ResearchQuestion}
Upon studying a broad range of works on false alarm mitigation, we call special attention to important open issues:
\begin{itemize}[leftmargin=*]
    \item \textit{Evaluation based on public benchmarks.} Most of the works evaluate their methods based on a local or custom dataset. Analyzing the methods' performance in terms of the false positives reduction ratio on a common dataset, like NAB or PyOD \cite{lavin2015evaluating,zhao2019pyod} can help to understand their usefulness and applicability.
    
    \item \textit{Evaluation based on common scoring mechanisms.} Results reported by many of the studies are often not clear indicators of a method's capabilities in different respects, as addressing the challenges mentioned in Section \ref{sec:CCADS}. Collecting measures and labeled datasets to evaluate each arbitrary method in all its respects is a necessity.
    
    \item {\it Uniformity.} Since most of the techniques use a custom range and format, comparing them is not straightforward. Thus, a uniform format to present anomaly scores and their ranking is required. Besides enhancing comparability, uniformity provides the possibility of ensembling several ADS methods.
    
    {\item Performance.} The resources required by the AD algorithm, including time and space complexity, are also  important characteristics for damage control, particularly in real-time applications. Most studies do however not report performance aspects. 
    
    \item {\it Addressing data drift.} Real-world stream datasets typically include a gradual or sharp shift over time. Methods should adjust themselves using incremental learning or other methodical approaches to address drifts and evolutions which happen in the target domain.
\end{itemize}

\section{Conclusions}
To the best of our knowledge, a comprehensive study on the critical steps of false alarm mitigation in the anomaly detection context is badly needed.
This paper provides an analytic review of methods found in the literature at the time of writing. 
Most of the studied methods focus on improving the behavior modeling phase, while the final scoring can immensely influence the system applicability. We have collected here the known strategies that can contribute to reduce the false alarm rate.

\textit{Predictive model improvement} focuses on extracting complex latent patterns from data; the \textit{anomaly scoring improvement} strategies take advantage of the rarity and probability values to score anomalies. The essential techniques of \textit{threshold computation} aim to overcome the difficulties of finding the best threshold to distinguish normal and anomalous cases. 
\textit{Post-hoc pruning} processes are another type of strategies to update the threshold related parameters based on the running system performance. {\it Collective analysis} strategies rescale assigned anomaly scores based on a collection of observations, like their correlation with each other. 
Also {\it sequence based scoring} approaches and their applications to mitigate false alarm rates by analyzing sub-sequences are briefly reviewed.

Despite the many existing approaches, further improvements are needed for mitigating false alarms so as to make anomaly detection practicable for real-world application domains, e.g.\ in the cyber arena, for threat detection and blocking increasingly sophisticated zero-day exploits. After all, even a very small false-alarm rate can mean an overwhelming absolute number of false positives---a notorious challenge which to overcome is still a long way to go.


\end{document}